\documentclass[10pt,twocolumn,letterpaper]{article}

\usepackage{cvpr}
\usepackage{times}
\usepackage{epsfig}
\usepackage{graphicx}
\usepackage{amsmath}
\usepackage{amssymb}

\usepackage{cases}
\usepackage[linesnumbered,ruled,vlined,lined,boxed,commentsnumbered]{algorithm2e}
\usepackage{subfigure}
\usepackage{indentfirst}
\usepackage{multirow}
\usepackage{boldline}

\usepackage{enumitem}
\setitemize{noitemsep,topsep=0pt,parsep=0pt,partopsep=0pt}

\usepackage[table]{xcolor}

\usepackage[pagebackref=true,breaklinks=true,letterpaper=true,colorlinks,bookmarks=false]{hyperref}

\cvprfinalcopy % *** Uncomment this line for the final submission

\def\cvprPaperID{1502} % *** Enter the CVPR Paper ID here

\ifcvprfinal\fi
\begin{document}

%%%%%%%%% TITLE
\title{Large Scale Fine-Grained Categorization and Domain-Specific Transfer Learning}

\author{Yin Cui$^{1,2}$\thanks{Work done during internship at Google Research.}\hspace{15pt}Yang Song$^{3}$\hspace{15pt}Chen Sun$^{3}$\hspace{15pt}Andrew Howard$^{3}$\hspace{15pt}Serge Belongie$^{1,2}$\\
$^{1}$Department of Computer Science, Cornell University\hspace{20pt}$^{2}$Cornell Tech\hspace{20pt}$^{3}$Google Research}

\maketitle
%\thispagestyle{empty}

%%%%%%%%%%%%%%%%%% ABSTRACT
\begin{abstract}
Transferring the knowledge learned from large scale datasets (\eg,\ ImageNet) via fine-tuning offers an effective solution for domain-specific fine-grained visual categorization (FGVC) tasks (\eg,\ recognizing bird species or car make \& model).
In such scenarios, data annotation often calls for specialized domain knowledge and thus is difficult to scale.
In this work, we first tackle a problem in large scale FGVC. 
Our method won first place in iNaturalist 2017 large scale species classification challenge.
Central to the success of our approach is a training scheme that uses higher image resolution and deals with the long-tailed distribution of training data.
Next, we study transfer learning via fine-tuning from large scale datasets to small scale, domain-specific FGVC datasets.
We propose a measure to estimate domain similarity via Earth Mover's Distance and demonstrate that transfer learning benefits from pre-training on a source domain that is similar to the target domain by this measure. 
Our proposed transfer learning outperforms ImageNet pre-training and obtains state-of-the-art results on multiple commonly used FGVC datasets.
\end{abstract}

%%%%%%%%%%%%%%%%%% Introduction
\section{Introduction}

Fine-grained visual categorization (FGVC) aims to distinguish subordinate visual categories. 
Examples include recognizing natural categories such as species of birds~\cite{cub200, nabirds}, dogs~\cite{stanford_dog} and plants~\cite{flower_102, urban_tree}; or man-made categories such as car make \& model~\cite{stanford_car, CompCars}.
A successful FGVC model should be able to discriminate categories with subtle differences, which presents formidable challenges for the model design yet also provides insights to a wide range of applications such as rich image captioning~\cite{anne2016deep}, image generation~\cite{CVAE-GAN}, and machine teaching~\cite{becoming_expert, mac2018teaching}.

\begin{figure}[t]
\begin{center}
\includegraphics[width=\columnwidth]{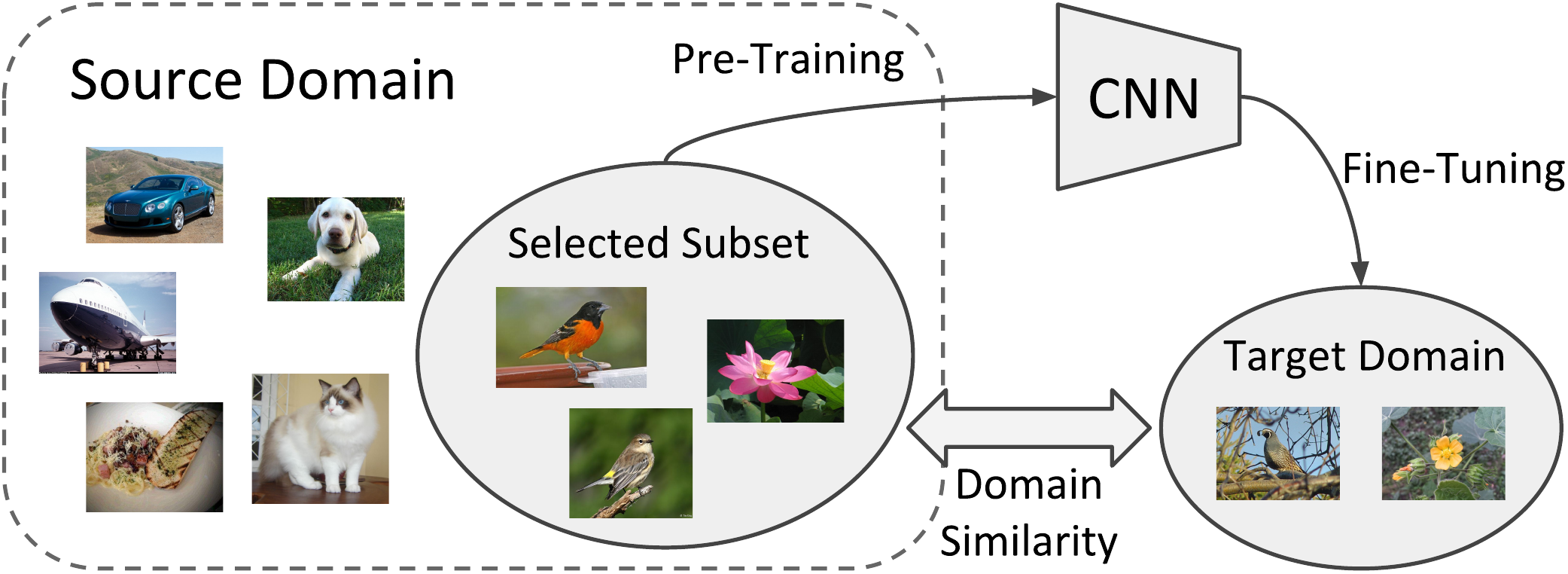}
\end{center}
\caption{Overview of the proposed transfer learning scheme.
% Given the target domain of interest, we select a subset of categories from the source domain based on a domain similarity measure.
Given the target domain of interest, we pre-train a CNN on the selected subset from the source domain based on the proposed domain similarity measure, and then fine-tune on the target domain.}
\label{fig:overview}
% \vspace{-2mm}
\end{figure}

Recent advances on Convolutional Neural Networks (CNNs) for visual recognition~\cite{alexnet, vggnet, googlenet, resnet} have fueled remarkable progress on FGVC~\cite{bilinearcnn, kernel_pooling, multi-attention_fgvc}.
In general, to achieve reasonably good performance with CNNs, one needs to train networks with vast amounts of supervised data.
However, collecting a labeled fine-grained dataset often requires expert-level domain knowledge and therefore is difficult to scale.
As a result, commonly used FGVC datasets~\cite{cub200, stanford_dog, stanford_car} are relatively small, typically containing around 10k of labeled training images.
In such a scenario, fine-tuning the networks that are pre-trained on large scale datasets such as ImageNet~\cite{imagenet} is often adopted. 

This common setup poses two questions:
%that require our deeper understanding:
1) What are the important factors to achieve good performance on large scale FGVC? 
Although other large scale generic visual datasets like ImageNet contain some fine-grained categories, their images are usually iconic web images that contain objects in the center with similar scale and simple backgrounds.
With the limited availability of large scale FGVC datasets, how to design models that perform well on large scale non-iconic images with fine-grained categories remains an underdeveloped area.
2) How does one effectively conduct transfer learning, by first training the network on a large scale dataset and then fine-tuning it on domain-specific fine-grained datasets?
Modern FGVC methods overwhelmingly use ImageNet pre-trained networks for fine-tuning.
Given the fact that the target fine-grained domain is known, can we do better than ImageNet?

This paper aims to answer the two aforementioned problems, with the recently introduced iNaturalist 2017 large scale fine-grained dataset (iNat)~\cite{inaturalist}.
iNat contains 675,170 training and validation images from 5,089 fine-grained categories. All images were captured in natural conditions with varied object scales and backgrounds.
Therefore, iNat offers a great opportunity to investigate key factors behind training CNNs that perform well on large scale FGVC. 
In addition, along with ImageNet, iNat enables us to study the transfer of knowledge learned on large scale datasets to small scale fine-grained domains.

In this work, we first propose a training scheme for large scale fine-grained categorization, achieving top performance on iNat.
% Specifically, we focus on exploring what benefits we can get from input images with higher resolution and how to deal with long-tailed distributed training data.
Unlike ImageNet, images in iNat have much higher resolutions and a wide range of object scales.
We show in Sec.\ \ref{sec:image_resolution} that performance on iNat can be improved significantly with higher input image resolution.
Another issue we address in this paper is the long-tailed distribution, where a few categories have most of the images~\cite{long-tail_object, van2017devil}.
To deal with this, we present a simple yet effective approach.
The idea is to learn good features from a large amount of training data and then fine-tune on a more evenly-distributed subset to balance the network's efforts among all categories and transfer the learned features.
Our experimental results, shown in Sec.\ \ref{sec:long-tailed}, reveal that we can greatly improve the under-represented categories and achieve better overall performance.

Secondly, we study how to transfer from knowledge learned on large scale datasets to small scale fine-grained domains.
Datasets are often biased in terms of their statistics on content and style~\cite{torralba2011unbiased}.
% Because of the biases, we found better transfer learning performance when fine-tuning the network trained on the large scale source dataset that is more visually similar to the small scale fine-grained target dataset.
On CUB200 Birds~\cite{cub200}, iNat pre-trained networks perform much better than ImageNet pre-trained ones; whereas on Stanford-Dogs~\cite{stanford_dog}, ImageNet pre-trained networks yield better performance.
This is because there are more visually similar bird categories in iNat and dog categories in ImageNet.
% This issue cannot be solved by transferring from a combined dataset of iNat and ImageNet.
In light of this, we propose a novel way to measure the visual similarity between source and target domains based on image-level visual similarity with Earth Mover's Distance.
By fine-tuning the networks trained on selected subsets based on our proposed domain similarity, we achieve better transfer learning than ImageNet pre-training and state-of-the-art results on commonly used fine-grained datasets.
% We show that better transfer learning performance can be obtained from the selected source domain that is similar to the given target domain.
Fig.\ \ref{fig:overview} gives an overview of the proposed training scheme.

% Our key contributions can be summarized as follows: 
% \begin{itemize}
% \item We propose a training scheme for large scale fine-grained categorization focused on leveraging input images with higher resolution and dealing with long-tailed distribution of the data. We conduct comprehensive experiments and our model achieves top performance on the challenging iNaturalist dataset~\cite{inaturalist}.
% \item We propose a novel way to quantify visual similarity between source and target domains and study the effectiveness of transfer learning with respect to domain similarity.
% By fine-tuning the networks trained on selected subsets based on our proposed domain similarity, we achieve better transfer learning than ImageNet pre-training and state-of-the-art results on commonly used fine-grained datasets.
% \end{itemize}
We believe our study on large scale FGVC and domain-specific transfer learning could offer useful guidelines for researchers working on similar problems.

%%%%%%%%%%%%%%%%%% Related Work
\section{Related Work}

\textbf{Fine-Grained Visual Categorization (FGVC)}.
Recent FGVC methods typically incorporate useful fine-grained information into a CNN and train the network end-to-end.
Notably, second order bilinear feature interactions was shown to be very effective~\cite{bilinearcnn}.
This idea was later extended to compact bilinear pooling~\cite{cbp}, and then higher order interactions~\cite{kernel_pooling, cai2017higher-order, simon2017generalized}.
To capture subtle visual differences, visual attention~\cite{two-level_attention, fu2017look, multi-attention_fgvc} and deep metric learning~\cite{facenet, cui2016fine} are often used.
Beyond pixels, we also leverage other information including parts~\cite{zhang2014part, pose_normlized_net, zhang2015fine}, attributes~\cite{fine-grained_attributes, gebru2017fine}, human interactions~\cite{branson2010visual, deng2016leveraging} and text descriptions~\cite{fine-grained_description, he2017fine}.
To deal with the lack of training data in FGVC, additional web images can be collected to augment the original dataset~\cite{cui2016fine, krause2016unreasonable, xu2016webly, gebru2017fine}.
Our approach differs from them by transferring the pre-trained network on existing large scale datasets without collecting new data.

Using high-resolution images for FGVC has became increasingly popular~\cite{STN,bilinearcnn}.
There is also a similar trend in ImageNet visual recognition, from originally $224 \times 224$ in AlexNet~\cite{alexnet} to $331 \times 331$ in recently proposed NASNet~\cite{nasnet}.
% There is also work showing fine-tuning the network trained on high-resolution FGVC images greatly improves the performance on low-resolution ones~\cite{peng2016low-res}.
However, no prior work has systematically studied the effect of image resolution on large scale fine-grained datasets as we do in this paper.

How to deal with long-tailed distribution is an important problem in real world data~\cite{long-tail_object, van2017devil}.
However, it is a rather unexplored area mainly because commonly used benchmark datasets are pre-processed to be close-to evenly distributed~\cite{imagenet, coco}.
Van Horn \etal~\cite{van2017devil} pointed out that the performance of tail categories are much poorer than head categories that have enough training data.
% And this problem cannot be easily solved by collecting more data for the head categories.
% \cite{long-tail_object} proposed a complex iterative framework to share and mix features across subcategories.
We present a simple two-step training scheme to deal with long-tailed distribution that works well in practice.

\textbf{Transfer Learning}.
% Since the ground breaking work of using Convolutional Neural Networks (CNNs) on ImageNet~\cite{alexnet}, 
Convolutional Neural Networks (CNNs) trained on ImageNet have been widely used for transfer learning, either by directly using the pre-trained network as a feature extractor~\cite{cnn_off_the_shelf, decaf, places}, or fine-tuning the network~\cite{rcnn, transfer_cvpr14}.
Due to the remarkable success of using pre-trained CNNs for transfer learning, extensive efforts have been made on understanding transfer learning~\cite{how_transferable, azizpour2016factors, what_makes_imagenet_transfer, revisiting_data}.
% Yosinski \etal~\cite{how_transferable} examined the transferability of features from each layer of a network, which reveals their generality or specificity.
% Azizpour \etal~\cite{azizpour2016factors} and Huh \etal~\cite{what_makes_imagenet_transfer} studied different factors in transfer learning, including network depth / width, number of images, number of categories, fine vs.\ coarse categories and so on.
% Chen \etal~\cite{revisiting_data} trained CNNs on a much larger and more generic JFT dataset~\cite{jft} and showed that transfer learning performance for object detection increases logarithmically as the number of data increases.
In particular, some prior work loosely demonstrated the connection between transfer learning and domain similarity.
For example, transfer learning between two random splits is easier than natural / man-made object splits in ImageNet~\cite{how_transferable}; manually adding $512$ additional relevant categories from all available classes improve upon the commonly used $1000$ ImageNet classes on PASCAL VOC~\cite{voc}; transferring from a combined ImageNet and Places dataset yields better results on a list of visual recognition tasks~\cite{places}.
Azizpour \etal~\cite{azizpour2016factors} conducted a useful study on a list of transfer learning tasks that have different similarity with the original ImageNet classification task (\eg, image classification is considered to be more similar than instance retrieval, \etc).
Our major differences between their work are two-fold:
Firstly, we provide a way to quantify the similarity between source and target domain and then choose a more similar subset from source domain for better transfer learning.
Secondly, they all use pre-trained CNNs as feature extractors and only train either the last layer or use a linear SVM on the extracted features, whereas we fine-tune all the layers of the network.

% 3. Selecting Source Domain based on Similarity in NLP?? 
% Learning to select data for transfer learning with Bayesian Optimization.
% Knowledge Adaptation: Teaching to Adapt.
% Data Selection Strategies for Multi-Domain Sentiment Analysis.

% 4. Meta-learning??
% Recently, there are some work on meta-learning, especially how to get a good initialization for better learning (faster convergence and better performance). 
% Our method can be also seen as an instance of meta-learning to learn a better initialization from data.
% Model-Agnostic Meta-Learning. 
% Meta-SGD https://arxiv.org/abs/1707.09835

%%%%%%%%%%%%%%%%%% large scale FGVC
\section{Large Scale Fine-Grained Categorization}
\label{sec:lsfgvc}

In this section, we present our training scheme that achieves top performance on the challenging iNaturalist 2017 dataset, especially focusing on using higher image resolution and dealing with long-tailed distribution.

\subsection{The Effect of Image Resolution}
\label{sec:image_resolution}

When training a CNN, for the ease of network design and training in batches, the input image is usually pre-processed to be square with a certain size.
Each network architecture usually has a default input size.
For example, AlexNet~\cite{alexnet} and VGGNet~\cite{vggnet} take the default input size of $224 \times 224$ and this default input size cannot be easily changed because the fully-connected layer after convolutions requires a fixed size feature map.
More recent networks including ResNet~\cite{resnet} and Inception~\cite{googlenet, inception-v3, inception-v4} are fully convolutional, with a global average pooling layer right after convolutions.
This design enables the network to take input images with arbitrary sizes.
Images with different resolution induce feature maps of different down-sampled sizes within the network.

Input images with higher resolutions usually contain richer information and subtle details that are important to visual recognition, especially for FGVC.
Therefore, in general, higher resolution input image yields better performance.
For networks optimized on ImageNet, there is a trend of using input images with higher resolution for modern networks: from originally $224 \times 224$ in AlexNet~\cite{alexnet} to $331 \times 331$ in recently proposed NASNet~\cite{nasnet}, as shown in Table\ \ref{tab:input_size}.
However, most images from ImageNet have a resolution of $500 \times 375$ and contain objects of similar scales, limiting the benefits we can get from using higher resolution inputs.
We explore the effect of using a wide range of input image sizes from $299 \times 299$ to $560 \times 560$ in iNat dataset, showing greatly improved performance with higher resolution inputs.

\subsection{Long-Tailed Distribution}
\label{sec:long-tailed}

The statistics of real world images is long-tailed: a few categories are highly representative and have most of the images, whereas most categories are observed rarely with only a few images~\cite{long-tail_object, van2017devil}.
This is in stark contrast to the even image distribution in popular benchmark datasets such as ImageNet~\cite{imagenet}, COCO~\cite{coco} and CUB200~\cite{cub200}.

With highly imbalanced numbers of images across categories in iNaturalist dataset~\cite{inaturalist}, we observe poor performance on underrepresented tail categories.
We argue that this is mainly caused by two reasons: 
1) The lack of training data. Around 1,500 fine-grained categories in iNat training set have fewer than 30 images. 
%In such a scenario, a network is difficult to achieve good performace on these long-tailed categories.
2) The extreme class imbalance encountered during training: the ratio between the number of images in the largest class and the smallest one is about 435.
Without any re-sampling of the training images or re-weighting of the loss, categories with more images in the head will dominate those in the tail.
Since there is very little we can do for the first issue of lack of training data, we propose a simple and effective way to address the second issue of the class imbalance.

\begin{figure}[t]
\centering
\includegraphics[width=\columnwidth]{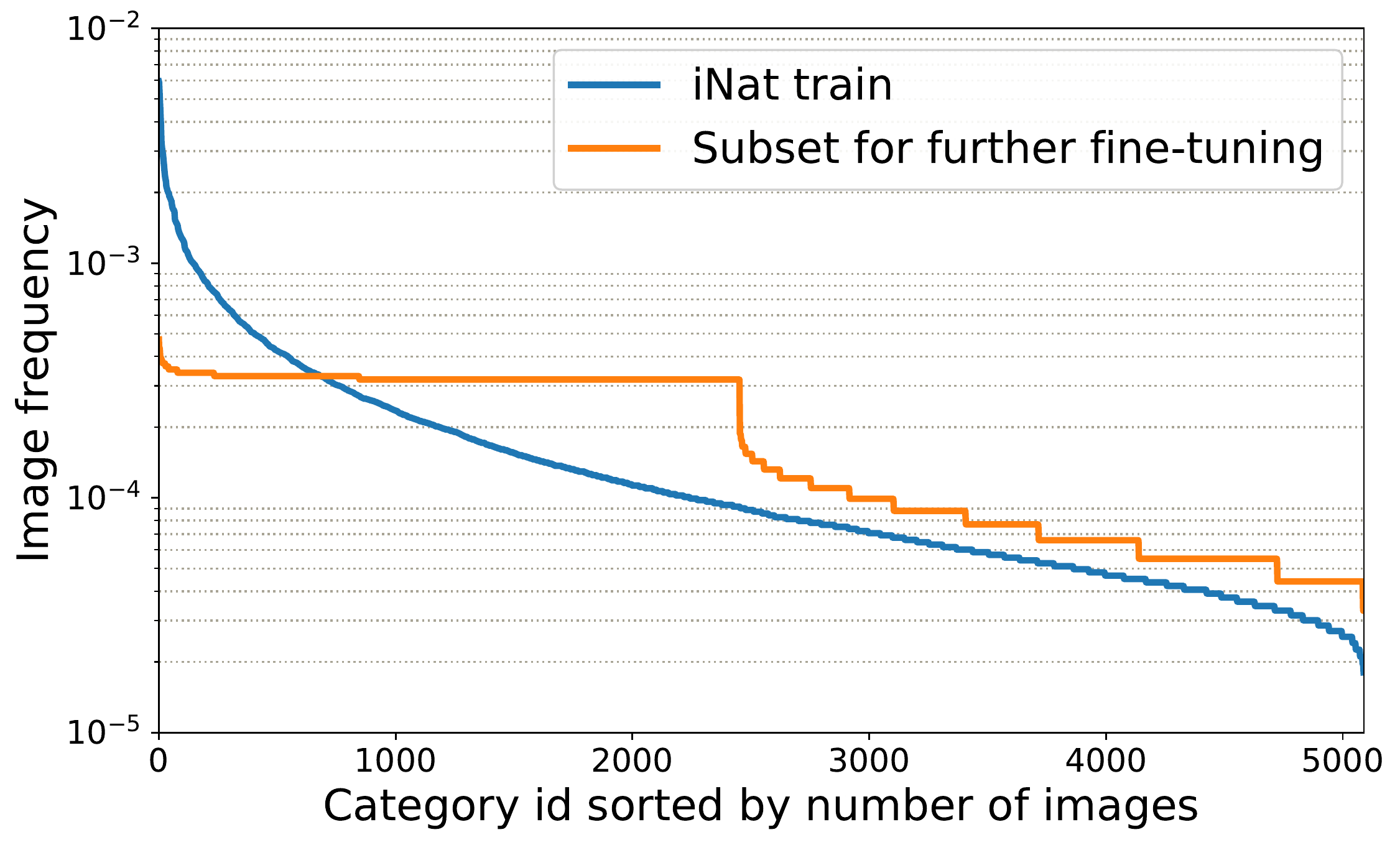}
\caption{The distribution of image frequency of each category in the whole training set we used in the first stage training and the selected subset we used in the second stage fine-tuning.}
\label{fig:long-tailed}
% \vspace{-2mm}
\end{figure}

The proposed training scheme has two stages.
In the first stage, we train the network as usual on the original imbalanced dataset.
With large number of training data from all categories, the network learns good feature representations.
Then, in the second stage, we fine-tune the network on a subset containing more balanced data with a small learning rate.
The idea is to slowly transfer the learned feature and let the network re-balance among all categories.
Fig.\ \ref{fig:long-tailed} shows the distribution of image frequency in iNat training set that we trained on in the first stage and the subset we used in the second stage, respectively.
Experiments in Sec.~\ref{sec:exp_inat} verify that the proposed strategy yields improved overall performance, especially for underrepresented tail categories.

\begin{table}[t]
\small
\begin{center}
\begin{tabular}{ |l|l| } 
\hline
Input Res. & Networks \\ \hline
$224 \times 224$ & AlexNet~\cite{alexnet}, VGGNet~\cite{vggnet}, ResNet~\cite{resnet} \\
$299 \times 299$ & Inception~\cite{googlenet, inception-v3, inception-v4} \\
$320 \times 320$ & ResNetv2~\cite{resnet-v2}, ResNeXt~\cite{resnext}, SENet~\cite{senet} \\
$331 \times 331$ & NASNet~\cite{nasnet} \\ \hline
\end{tabular}
\end{center}
\caption{Default input image resolution for different networks.
There is a trend of using input images with higher resolution for modern networks.
}
\label{tab:input_size}
% \vspace{-2mm}
\end{table}

%%%%%%%%%%%%%%%%%% Transfer Learning
\section{Transfer Learning}
\label{sec:transfer_learning}

This section describes transfer learning from the networks trained on large scale datasets to small scale fine-grained datasets.
We introduce a way to measure visual similarity between two domains and then show how to select a subset from source domain given the target domain.

\subsection{Domain Similarity}
\label{sec:domain_similarity}

Suppose we have a source domain $\mathcal{S}$ and a target domain $\mathcal{T}$.
We define the distance between two images $s \in \mathcal{S}$ and $t \in \mathcal{T}$ as the Euclidean distance between their feature representations:
\begin{equation}
\label{eqn:image_dist}
d(s, t) = \lVert g(s) - g(t) \rVert
\end{equation}
where $g(\cdot)$ denotes a feature extractor for an image.
To better capture the image similarity, the feature extractor $g(\cdot)$ needs to be capable of extracting high-level information from images in a generic, unbiased manner.
Therefore, in our experiments, we use $g(\cdot)$ as the features extracted from the penultimate layer of a ResNet-101 trained on the large scale JFT dataset~\cite{revisiting_data}.

In general, using more images yields better transfer learning performance.
For the sake of simplicity, in this study
%of domain similarity and transfer learning,
we ignore the effect of domain scale (number of images).
Specifically, we normalize the number of images in both source and target domain.
As studied by Chen \etal~\cite{revisiting_data}, transfer learning performance increases logarithmically with the amount of training data.
This suggests that the performance gain in transfer learning resulting from the use of more training data would be insignificant when we already have a large enough dataset (\eg, ImageNet).
Therefore, ignoring the domain scale is a reasonable assumption that simplifies the problem.
Our definition of domain similarity can be generalized to take domain scale into account by adding a scale factor, but we found ignoring the domain scale already works well in practice.
% and our associated experimental evaluation.
% This assumption is also verified in experiments.

Under this assumption, transfer learning can be viewed as moving a set of images from the source domain $\mathcal{S}$ to the target domain $\mathcal{T}$.
The work needed to be done by moving an image to another can be defined as their image distance in Eqn.\ \ref{eqn:image_dist}.
Then the distance between two domains can be defined as the least amount of total work needed.
This definition of domain similarity can be calculated by the Earth Mover's Distance (EMD)~\cite{emd, emd_2}.

\begin{figure}[t]
\begin{center}
\includegraphics[width=\columnwidth]{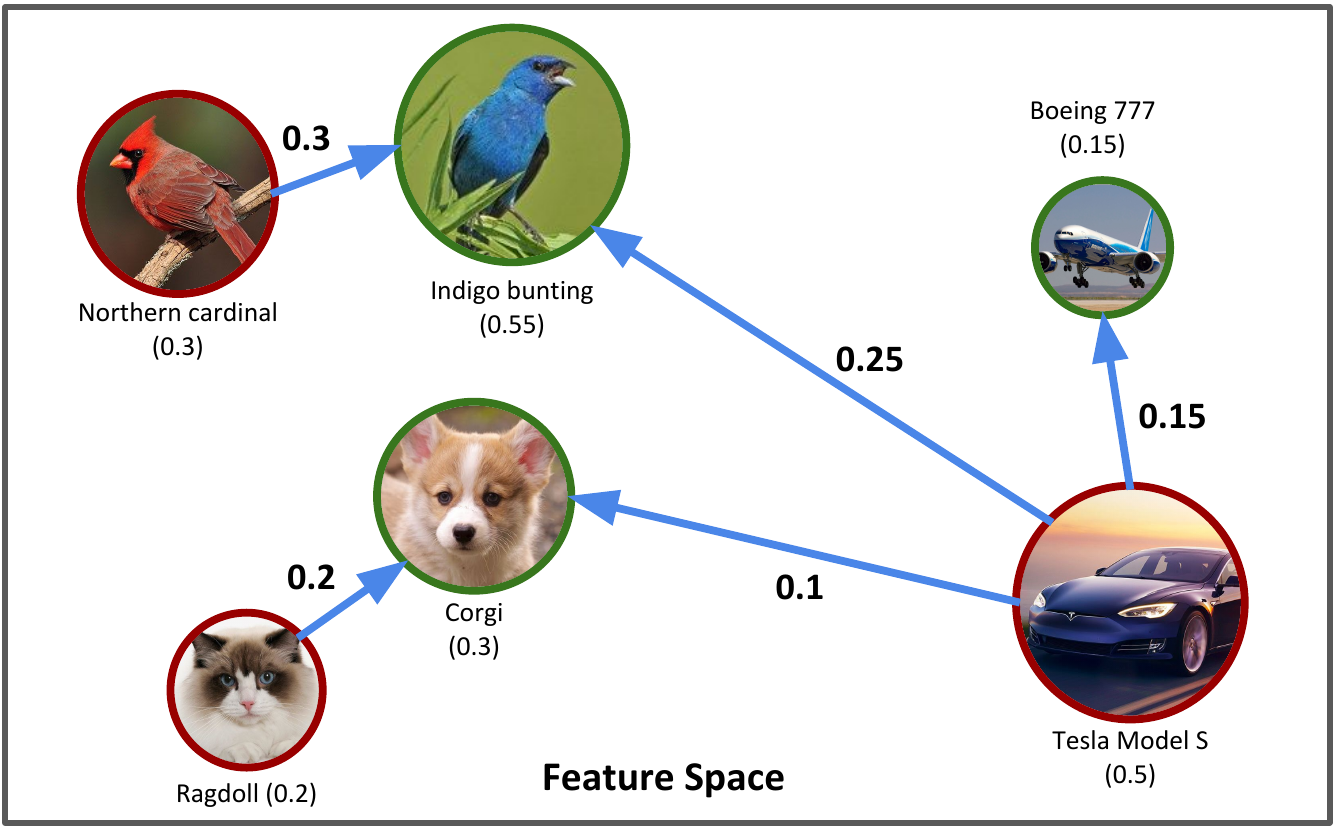}
\end{center}
\caption{The proposed domain similarity calculated by Earth Mover's Distance (EMD). Categories in source domain and target domain are represented by red and green circles. The size of the circle denotes the normalized number of images in that category. Blue arrows represent flows from source to target domain by solving EMD.}
\label{fig:emd}
% \vspace{-2mm}
\end{figure}

To make the computations more tractable, we further make an additional simplification to represent all image features in a category by the mean of their features.
Formally, we denote source domain as $\mathcal{S} = \{ (s_i, w_{s_i}) \}_{i=1}^m$ and target domain as $\mathcal{T} = \{ (t_j, w_{t_j}) \}_{j=1}^n$, where $s_i$ is the $i$-th category in $\mathcal{S}$ and $w_{s_i}$ is the normalized number of images in that category; similarly for $t_j$ and $w_{t_j}$ in $\mathcal{T}$.
$m$ and $n$ are the total number of categories in source domain $\mathcal{S}$ and target domain $\mathcal{T}$, respectively.
Since we normalize the number of images, we have $\sum_{i=1}^m w_{s_i} = \sum_{j=1}^n w_{t_j} = 1$.
$g(s_i)$ denotes the mean of image features in category $i$ from source domain, similarly for $g(t_j)$ in target domain.
Using the defined notations, the distance between $\mathcal{S}$ and $\mathcal{T}$ is defined as their Earth Mover's Distance (EMD):
\begin{equation}
\label{eqn:emd}
d(\mathcal{S}, \mathcal{T}) = \text{EMD}(\mathcal{S}, \mathcal{T}) = \frac{\sum_{i=1, j=1}^{m, n} f_{i,j} d_{i,j}}{\sum_{i=1, j=1}^{m, n} f_{i,j}}
\end{equation}
where $d_{i, j}=\lVert g(s_i)-g(t_j)\rVert$ and the optimal flow $f_{i, j}$  corresponds to the least amount of total work by solving the EMD optimization problem.
% \begin{equation*}
% \begin{aligned}
% & \underset{f_{i,j}}{\text{min}} & & \sum_{i=1}^m \sum_{j=1}^n f_{i,j} d_{i,j} \\
% & \text{subject to} & & f_{i,j} \geq 0 \\
% & & & \sum_{j=1}^n f_{i,j} \leq w_{s_i}, \sum_{i=1}^m f_{i,j} \leq w_{t_j} \\
% & & & \sum_{i=1}^m \sum_{j=1}^n f_{i,j} = \text{min} \Big\{ \sum_{i=1}^m w_{s_i}, \sum_{j=1}^n w_{t_j} \Big\}
% \end{aligned}
% \end{equation*}
Finally, the domain similarity is defined as:
\begin{equation}
\label{eqn:domain_similarity}
\text{sim}(\mathcal{S}, \mathcal{T}) = e^{-\gamma d(\mathcal{S}, \mathcal{T})}
\end{equation}
where $\gamma$ is set to 0.01 in all experiments.
Fig.\ \ref{fig:emd} illustrates calculating the proposed domain similarity by EMD.

\subsection{Source Domain Selection}
\label{sec:selection}

With the defined domain similarity in Eqn.\ \ref{eqn:emd}, we are able to select a subset from source domain that is more similar to target domains.
We use greedy selection strategy to incrementally include the most similar category in the source domain.
That is, for each category $s_i$ in source domain $\mathcal{S}$, we calculate its domain similarity with target domain by $\text{sim}(\{(s_i, 1)\}, \mathcal{T})$ as defined in Eqn.\ \ref{eqn:domain_similarity}.
Then top $k$ categories with highest domain similarities will be selected.
Notice that although this greedy way of selection has no guarantee on the optimality of the selected subset of size $k$ in terms of domain similarity, we found this simple strategy works well in practice.

% Details of the domain selection is shown in Alg.\ \ref{alg:domain_selection}.

% \begin{algorithm}[t]
% \label{alg:domain_selection}
% \KwIn{$\mathcal{S} = \{ (s_i, w_{s_i}) \}_{i=1}^m$, $\mathcal{T} = \{ (t_j, w_{t_j}) \}_{j=1}^n$, $k \leq m$}
% \KwOut{$\mathcal{D} \subset \mathcal{S}$, where $|\mathcal{D}| = k$}

% Initialization: $\mathcal{D} \leftarrow \emptyset$, d \leftarrow []\\
% \For{$i\leftarrow 1$ \KwTo $m$}{
% \text{EMD}(\{(s_i, 1)\}, \mathcal{T})\\
% }
% \caption{Source Domain selection}
% \end{algorithm}

%%%%%%%%%%%%%%%%%% Experiments
\section{Experiments}
\label{sec:experiments}

The proposed training scheme for large scale FGVC is evaluated on the recently proposed iNaturalist 2017 dataset (iNat)~\cite{inaturalist}.
We also evaluate the effectiveness of the our proposed transfer learning by using ImageNet and iNat as source domains, and 7 fine-grained categorization datasets as target domains.
Sec.\ \ref{sec:exp_setup} introduces experiment setup.
Experiment results on iNat and transfer learning are presented in Sec.\ \ref{sec:lsfgvc} and Sec.\ \ref{sec:exp_fgvc}, respectively.

\subsection{Experiment setup}
\label{sec:exp_setup}

\subsubsection{Datasets}
\label{sec:exp_datasets}

\textbf{iNaturalist}.
The iNatrualist 2017 dataset (iNat)~\cite{inaturalist} contains 675,170 training and validation images from 5,089 natural fine-grained categories.
Those categories belong to 13 super-categories including Plantae (Plant), Insecta (Insect), Aves (Bird), Mammalia (Mammal), and so on.
% The numbers of fine-grained categories and images per super-category are list in Table\ \ref{}.
The iNat dataset is highly imbalanced with dramatically different number of images per category.
For example, the largest super-category ``Plantae (Plant)" has 196,613 images from 2,101 categories; whereas the smallest super-category ``Protozoa" only has 381 images from 4 categories.
We combine the original split of training set and 90\% of the validation set as our training set (iNat train), and use the rest of 10\% validation set as our mini validation set (iNat minival), resulting in total of 665,473 training and 9,697 validation images.

% \begin{table}[t]
% \footnotesize
% \centering
% \begin{tabular}{ |l|r|r| } 
% \hline
% Super-class & \# class & \# image \\ \hline
% Plantae & 2,101 & 196,613 \\
% Insecta & 1,021 & 118,555 \\
% Aves & 964 & 235,521 \\
% Reptilia & 289 & 40,881 \\
% Mammalia & 186 & 32,823\\
% Fungi & 121 & 7,606 \\
% Amphibia & 115 & 17,703 \\
% Mollusca & 93 & 9,377\\
% Animalia & 77 & 6,590\\
% Arachnida & 56 & 5,959 \\
% Actinopterygii & 53 & 2,619 \\
% Chromista & 9 & 542 \\
% Protozoa & 4 & 381 \\\hline
% {\bf Total} & 5,089 & 675,170 \\\hline
% \end{tabular}
% \caption{}
% \label{tab:inat_dataset}
% \end{table}

\textbf{ImageNet}.
We use the ILSVRC 2012~\cite{ilsvrc} splits of 1,281,167 training (ImageNet train) and 50,000 validation (ImageNet val) images from 1,000 classes.

\textbf{Fine-Grained Visual Categorization}.
We evaluate our transfer learning approach on 7 fine-grained visual categorization datasets as target domains, which cover a wide range of FGVC tasks including natural categories like bird and flower and man-made categories such as aircraft. 
Table \ref{tab:fgvc_dataset} summarizes number of categories, together with number of images in their original training and validation splits.

\subsubsection{Network Architectures}
\label{sec:exp_network}

We use 3 types of network architectures: ResNet~\cite{resnet, resnet-v2}, Inception~\cite{googlenet, inception-v3, inception-v4} and SENet~\cite{senet}.

\textbf{Residual Network (ResNet)}.
Originally introduced by He \etal~\cite{resnet}, networks with residual connections greatly reduced the optimization difficulties and enabled the training of much deeper networks.
ResNets were later improved by pre-activation that uses identity mapping as the skip connection between residual modules~\cite{resnet-v2}.
We used the latest version of ResNets~\cite{resnet-v2} with 50, 101 and 152 layers.

\textbf{Inception}.
The Inception module was firstly proposed by Szegedy \etal in GoogleNet~\cite{googlenet} that was designed to be very efficient in terms of parameters and computations, while achieving state-of-the-art performance.
Inception module was then further optimized by using Batch Normalization~\cite{bn}, factorized convolution~\cite{inception-v3, inception-v4} and residual connections~\cite{inception-v4} as introduced in~\cite{resnet}.
We use Inception-v3~\cite{inception-v3}, Inception-v4 and Inception-ResNet-v2~\cite{inception-v4} as representatives for Inception networks in our experiments.

\textbf{Squeeze-and-Excitation (SE)}.
Recently proposed by Hu \etal~\cite{senet}, Sequeeze-and-Excitation (SE) modules achieved the best performance in ILSVRC 2017~\cite{ilsvrc}.
SE module squeezes responses from a feature map by spatial average pooling and then learns to re-scale each channel of the feature map.
Due to its simplicity in design, SE module can be used in almost any modern networks to boost the performance with little additional overhead.
We use Inception-v3 SE and Inception-ResNet-v2 SE as baselines.

For all network architectures, we follow strictly their original design but with the last linear classification layer replaced to match the number of categories in our datasets.

\begin{table}[t]
\small
\begin{center}
\begin{tabular}{ |l|r|r|r| } 
\hline
FGVC Dataset & \# class & \# train & \# val \\ \hline 
Flowers-102~\cite{flower_102} & 102 & 2,040 & 6,149 \\
CUB200 Birds~\cite{cub200}   & 200 & 5,994 & 5,794 \\
Aircraft~\cite{airplane}       & 100 & 6,667 & 3,333 \\
Stanford Cars~\cite{stanford_car}  & 196 & 8,144 & 8,041 \\
Stanford Dogs~\cite{stanford_dog}  & 120 & 12,000 & 8,580 \\
NABirds~\cite{nabirds}        & 555 & 23,929 & 24,633 \\
Food101~\cite{food101}        & 101 & 75,750 & 25,250 \\ \hline
\end{tabular}
\end{center}
\caption{We use 7 fine-grained visual categorization datasets to evaluate the proposed transfer learning method.}
\label{tab:fgvc_dataset}
% \vspace{-2mm}
\end{table}

\subsubsection{Implementation}
\label{sec:exp_implementation}

We used open-source Tensorflow~\cite{tensorflow} to implement and train all the models asynchronously on multiple NVIDIA Tesla K80 GPUs.
During training, the input image was randomly cropped from the original image and re-sized to the target input size with scale and aspect ratio augmentation~\cite{googlenet}.
We trained all networks using the RMSProp optimizer with momentum of 0.9, and the batch size of 32.
The initial learning rate was set to 0.045, with exponential decay of 0.94 after every 2 epochs, same as~\cite{googlenet}; for fine-tuning in transfer learning, the initial learning rate is lowered to 0.0045 with the learning rate decay of 0.94 after every 4 epochs.
We also used label smoothing as introduced in~\cite{inception-v3}.
During inference, the original image is center cropped and re-sized to the target input size.

% \begin{figure}[t]
% \centering
% \includegraphics[width=0.47\textwidth]{figures/input_size.pdf}
% \caption{Top-5 error rate on iNat minival with Inception-v3 trained and evaluated with various input sizes. Higher resolution inputs in both training and inference yield improved performance.}
% \label{fig:input_size}
% % \vspace{-2mm}
% \end{figure}

\begin{table}[t]
\small
\begin{center}
\begin{tabular}{ l|r|r|r }
\hline
& \textbf{Inc-v3 299} & \textbf{Inc-v3 448} & \textbf{Inc-v3 560} \\ \hline
Top-1 (\%) & 29.93 & 26.51 & 25.37 \\
Top-5 (\%) & 10.61 & 9.02 & 8.56 \\ \hline
\end{tabular}
\end{center}
\caption{Top-5 error rate on iNat minival using Inception-v3 with various input sizes. Higher input size yield better performance.}
\label{tab:input_size}
% \vspace{-2mm}
\end{table}

\subsection{Large Scale Fine-Grained Visual Recognition}
\label{sec:exp_inat}

To verify the proposed learning scheme for large scale fine-grained categorization, we conduct extensive experiments on iNaturalist 2017 dataset.
For better performance, we fine-tune from ImageNet pre-trained networks.
If training from scratch on iNat, the top-5 error rate is $\approx 1\%$ worse.

We train Inception-v3 with 3 different input resolutions (299, 448 and 560).
The effect of image resolution is presented in Table~\ref{tab:input_size}.
From the table, we can see that using higher input resolutions achieve better performance on iNat.
% Notably, under the same inference resolution, networks trained with higher resolutions perform better, indicating the network learns better discriminative knowledge with higher image resolution and those knowledge can be transferred to lower resolution images.

\begin{figure}[t]
\centering
\includegraphics[width=\columnwidth]{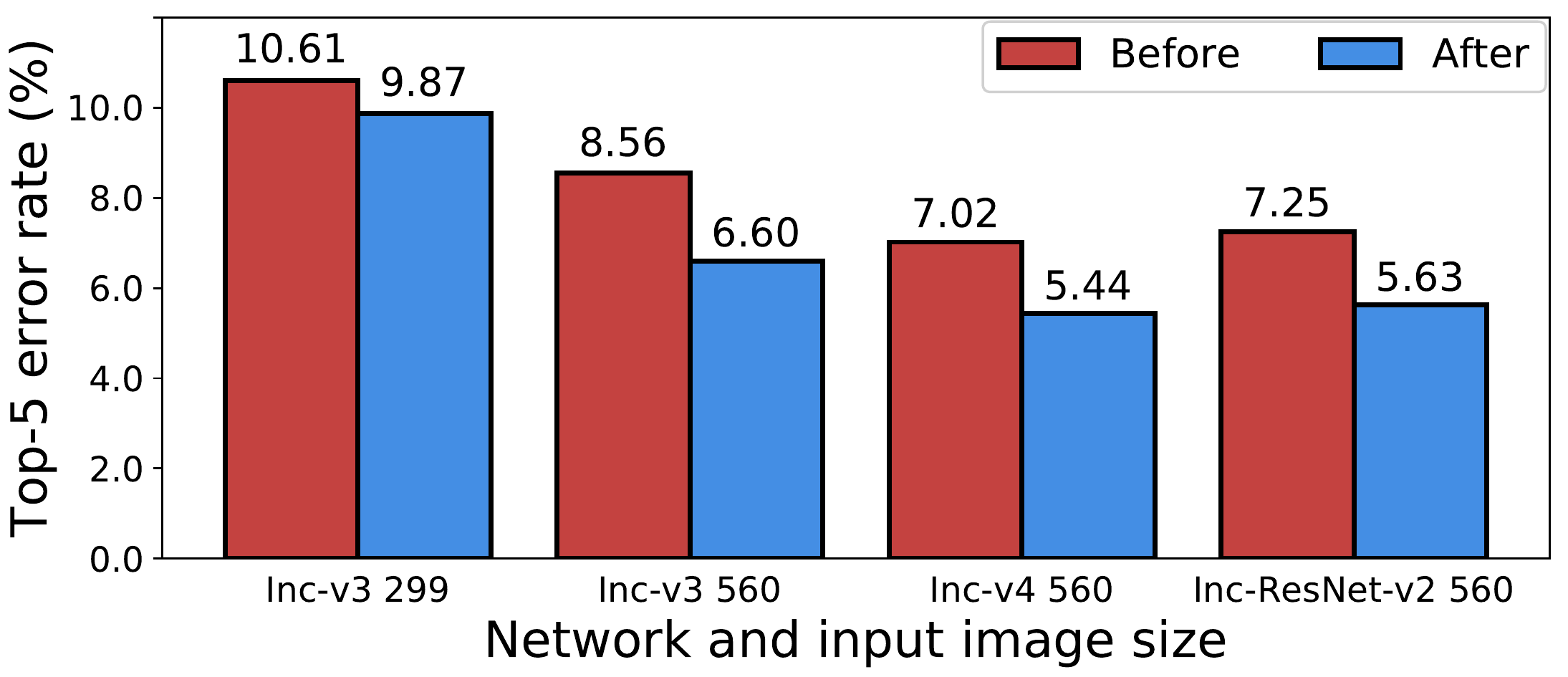}
\caption{Top-5 error rate on iNat minival before and after fine-tuning on a more balanced subset.
This simple strategy improves the performance on long-tailed iNat dataset.}
\label{fig:fine_tune}
% \vspace{-2mm}
\end{figure}

The evaluation of our proposed fine-tuning scheme for dealing with long-tailed distribution is presented in Fig.\ \ref{fig:fine_tune}.
Better performance can be obtained by further fine-tuning on a more balanced subset with small learning rate ($10^{-6}$ in our experiments).
Table \ref{tab:long-tailed} shows performance improvements on head and tail categories with fine-tuning.
Improvements on head categories with $\geq 100$ training images are 1.95\% of top-1 and 0.92\% of top-5; whereas on tail categories with $< 100$ training images, the improvements are 5.74\% of top-1 and 2.71\% of top-5.
These results verify that the proposed fine-tuning scheme greatly improves the performance on underrepresented tail categories.

Table \ref{tab:inat} presents the detailed performance breakdown of our winning entry in the iNaturalist 2017 challenge~\cite{inatchallenge}.
Using higher image resolution and further fine-tuning on a more balanced subset are the key to our success.

\subsection{Domain Similarity and Transfer Learning}
\label{sec:exp_fgvc}

We evaluate the proposed transfer learning method by pre-training the network on source domain \textit{from scratch}, and then fine-tune on target domains for fine-grained visual categorization.
Other than training separately on ImageNet and iNat, we also train networks on a combined ImageNet + iNat dataset that contains 1,946,640 training images from 6,089 categories (\ie, 1,000 from ImageNet and 5,089 from iNat).
We use input size of $299 \times 299$ for all networks.
Table \ref{tab:joint_training} shows the pre-training performance evaluated on ImageNet val and iNat minival.
Notably, a single network trained on the combined ImageNet + iNat dataset achieves competitive performance compared with two models trained separately. In general, combined training is better than training separately in the case of Inception and Inception SE, but worse in the case of ResNet.

\begin{figure*}[t!]
\begin{center}
\includegraphics[width=\textwidth]{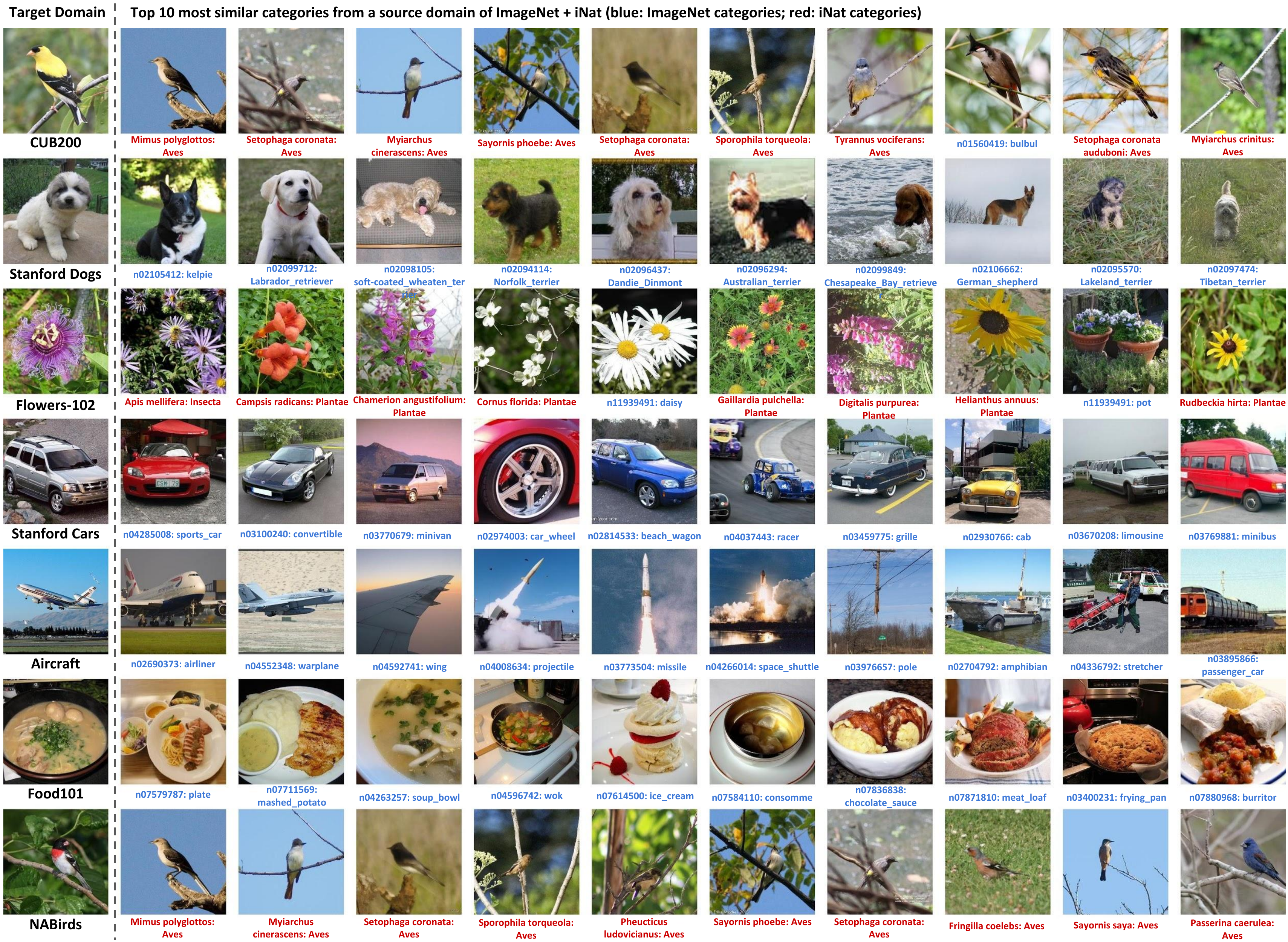}
\end{center}
\caption{Examples showing top 10 most similar categories in the combined ImageNet + iNat for each FGVC dataset, calculated with our proposed domain similarity. The left column represents 7 FGVC target domains, each by a randomly chosen image from the dataset. Each row shows top 10 most similar categories in ImageNet + iNat for a specific FGVC target domain. We represent a category by one randomly chosen image from that category. ImageNet categories are marked in blue, whereas iNat categories are in red.}
\label{fig:ds_examples}
% \vspace{-2mm}
\end{figure*}

Based on the proposed domain selection strategy defined in Sec.\ \ref{sec:selection}, we select the following two subsets from the combined ImageNet + iNat dataset:
\textbf{Subset A} was chosen by including top 200 ImageNet + iNat categories for each of the 7 FGVC dataset.
Removing duplicated categories resulted in a source domain containing 832 categories.
\textbf{Subset B} was selected by adding most similar 400 categories for CUB200, NABirds, top 100 categories for Stanford Dogs and top 50 categories for Stanford Cars and Aircraft, which gave us 585 categories in total.
Fig.\ \ref{fig:ds_examples} shows top 10 most similar categories in ImageNet + iNat for all FGVC datasets calculated by our proposed domain similarity.
It's clear to see that for CUB200, Flowers-102 and NABirds, most similar categories are from iNat; whereas for Stanford Dogs, Stanford Cars, Aircraft and Food101, most similar categories are from ImageNet.
This indicates the strong dataset bias in both ImageNet and iNat.

The transfer learning performance by fine-tuning an Inception-v3 on fine-grained datasets are presented in Table \ref{tab:fgvc}.
We can see that both ImageNet and iNat are highly biased, achieving dramatically different transfer learning performance on target datasets.
Interestingly, when we transfer networks trained on the combined ImageNet + iNat dataset, performance are in-between ImageNet and iNat pre-training, indicating that we cannot achieve good performance on target domains by simply using a larger scale, combined source domain.

\begin{table}[t]
\small
\begin{center}
\begin{tabular}{ l|r|r|r|r }
\hline
& \multicolumn{2}{c|}{\textbf{Before FT}} & \multicolumn{2}{c}{\textbf{After FT}} \\ \cline{2-5}
& \textbf{Top-1} & \textbf{Top-5} & \textbf{Top-1} & \textbf{Top-5} \\ \hline
Head: $\geq 100$ imgs & 19.28 & 5.79 & 17.33 & 4.87 \\
Tail: $< 100$ imgs & 29.89 & 9.12 & 24.15 & 6.41 \\ \hline
\end{tabular}
\end{center}
\caption{Top-1 and top-5 error rates (\%) on iNat minival for Inception-v4 560. The proposed fine-tuning scheme greatly improves the performance on underrepresented tail categories.}
\label{tab:long-tailed}
% \vspace{-2mm}
\end{table}

\begin{table}[t]
\small
\begin{center}
\begin{tabular}{ l|c|c }\hline 
{\bf Network} & {\bf Top-1 (\%)} & {\bf Top-5 (\%)} \\ \hline
Inc-v3 299    & 29.9 & 10.6 \\
Inc-v3 560    & 25.4 (+ 4.5) & 8.6 (+ 2.0) \\
Inc-v3 560 FT & 22.7 (+ 2.7) & 6.6 (+ 2.0) \\
Inc-v4 560 FT & 20.8 (+ 1.9) & 5.4 (+ 1.2) \\
Inc-v4 560 FT 12-crop & 19.2 (+ 1.6) & 4.7 (+ 0.7) \\
Ensemble     & 18.1 (+ 1.1) & 4.1 (+ 0.6) \\ \hline 
\end{tabular}
\end{center}
\caption{Performance improvements on iNat minival. The number inside the brackets indicates the improvement over the model in the previous row. FT denotes using the proposed fine-tuning to deal with long-tailed distribution. Ensemble contains two models: Inc-v4 560 FT and Inc-ResNet-v2 560 FT with 12-crop.}
\label{tab:inat}
% \vspace{-2mm}
\end{table}

\begin{table*}[t]
\small
\begin{center}
\begin{tabular}{ lcccccccccc }
\hline
\multirow{3}{*}{} & \multicolumn{6}{|c}{ImageNet val} & \multicolumn{4}{|c}{iNaturalist minival} \\
\cline{2-11}
 & \multicolumn{2}{|c}{Original} & \multicolumn{2}{|c}{Separate Train} & \multicolumn{2}{|c}{Combined Train} & \multicolumn{2}{|c}{Separate Train} & \multicolumn{2}{|c}{Combined Train} \\
\cline{2-11}
 & \multicolumn{1}{|c}{top-1} & \multicolumn{1}{|c}{top-5} & \multicolumn{1}{|c}{top-1} & \multicolumn{1}{|c}{top-5} & \multicolumn{1}{|c}{top-1} & \multicolumn{1}{|c}{top-5} & \multicolumn{1}{|c}{top-1} & \multicolumn{1}{|c}{top-5} & \multicolumn{1}{|c}{top-1} & \multicolumn{1}{|c}{top-5} \\ \hline
\multicolumn{1}{l}{ResNet-50~\cite{resnet,resnet-v2}} & \multicolumn{1}{|r}{24.70} & \multicolumn{1}{|r}{7.80} & \multicolumn{1}{|r}{\textbf{24.33}} & \multicolumn{1}{|r}{\textbf{7.61}} & \multicolumn{1}{|r}{25.23} & \multicolumn{1}{|r}{8.06} & \multicolumn{1}{|r}{\textbf{36.23}} & \multicolumn{1}{|r}{\textbf{15.67}} & \multicolumn{1}{|r}{36.93} & \multicolumn{1}{|r}{16.49} \\
\multicolumn{1}{l}{ResNet-101~\cite{resnet,resnet-v2}} & \multicolumn{1}{|r}{23.60} & \multicolumn{1}{|r}{7.10} & \multicolumn{1}{|r}{\textbf{23.08}} & \multicolumn{1}{|r}{7.09} & \multicolumn{1}{|r}{23.39} & \multicolumn{1}{|r}{\textbf{7.06}} & \multicolumn{1}{|r}{34.15} & \multicolumn{1}{|r}{14.58} & \multicolumn{1}{|r}{\textbf{33.97}} & \multicolumn{1}{|r}{\textbf{14.53}} \\
\multicolumn{1}{l}{ResNet-152~\cite{resnet,resnet-v2}} & \multicolumn{1}{|r}{23.00} & \multicolumn{1}{|r}{6.70} & \multicolumn{1}{|r}{\textbf{22.34}} & \multicolumn{1}{|r}{6.81} & \multicolumn{1}{|r}{22.59} & \multicolumn{1}{|r}{\textbf{6.64}} & \multicolumn{1}{|r}{\textbf{31.04}} & \multicolumn{1}{|r}{\textbf{12.52}} & \multicolumn{1}{|r}{32.58} & \multicolumn{1}{|r}{13.20} \\ \hline \hline
\multicolumn{1}{l}{Inception-v3~\cite{inception-v3}} & \multicolumn{1}{|r}{21.20} & \multicolumn{1}{|r}{5.60} & \multicolumn{1}{|r}{21.73} & \multicolumn{1}{|r}{5.97} & \multicolumn{1}{|r}{\textbf{21.52}} & \multicolumn{1}{|r}{\textbf{5.87}} & \multicolumn{1}{|r}{31.18} & \multicolumn{1}{|r}{11.90} & \multicolumn{1}{|r}{\textbf{30.29}} & \multicolumn{1}{|r}{\textbf{11.10}} \\
\multicolumn{1}{l}{Inception-ResNet-v2~\cite{inception-v4}} & \multicolumn{1}{|r}{$19.90^*$} & \multicolumn{1}{|r}{$4.90^*$} & \multicolumn{1}{|r}{20.33} & \multicolumn{1}{|r}{\textbf{5.16}} & \multicolumn{1}{|r}{\textbf{20.20}} & \multicolumn{1}{|r}{5.18} & \multicolumn{1}{|r}{\textbf{27.53}} & \multicolumn{1}{|r}{9.87} & \multicolumn{1}{|r}{27.78} & \multicolumn{1}{|r}{\textbf{9.12}} \\ \hline \hline
\multicolumn{1}{l}{Inception-v3 SE~\cite{senet}} & \multicolumn{1}{|r}{-} & \multicolumn{1}{|r}{-}  & \multicolumn{1}{|r}{20.98} & \multicolumn{1}{|r}{5.76} & \multicolumn{1}{|r}{\textbf{20.75}} & \multicolumn{1}{|r}{\textbf{5.69}} & \multicolumn{1}{|r}{30.15} & \multicolumn{1}{|r}{11.69} & \multicolumn{1}{|r}{\textbf{29.79}} & \multicolumn{1}{|r}{\textbf{10.64}} \\
\multicolumn{1}{l}{Inception-ResNet-v2 SE~\cite{senet}} & \multicolumn{1}{|r}{19.80} & \multicolumn{1}{|r}{4.79} & \multicolumn{1}{|r}{19.77} & \multicolumn{1}{|r}{4.79} & \multicolumn{1}{|r}{\textbf{19.56}} & \multicolumn{1}{|r}{\textbf{4.61}} & \multicolumn{1}{|r}{27.30} & \multicolumn{1}{|r}{9.61} & \multicolumn{1}{|r}{\textbf{26.01}} & \multicolumn{1}{|r}{\textbf{8.18}} \\ \hline
\end{tabular}
\end{center}
\caption{Pre-training performance on different source domains. Networks trained on the combined ImageNet + iNat dataset with 6,089 classes achieve competitive performance on both ImageNet and iNat compared with networks trained separately on each dataset. $*$ indicates the model was evaluated on the non-blacklisted subset of ImageNet validation set that may slightly improve the performance.}
\label{tab:joint_training}
% \vspace{-2mm}
\end{table*}

\begin{table*}[t]
\small
\begin{center}
\begin{tabular}{ l|c|c|c|c|c|c|c }
\hline
& CUB200 & Stanford Dogs & Flowers-102 & Stanford Cars & Aircraft & Food101 & NABirds \\ \hline
ImageNet & \cellcolor{red!25}82.84 & 84.19 & \cellcolor{red!25}96.26 & 91.31 & 85.49 & 88.65 & \cellcolor{red!25}82.01 \\
iNat & \cellcolor{green!25}89.26 & \cellcolor{red!25}78.46 & \cellcolor{green!25}97.64 & \cellcolor{red!25}88.31 & \cellcolor{red!25}82.61 & 88.80 & \cellcolor{green!25}87.91 \\
ImageNet + iNat & 85.84 & 82.36 & 97.07 & 91.38 & 85.21 & 88.45 & 83.98 \\
Subset A (832-class) & 86.37 & 84.69 & \cellcolor{green!25}97.65 & \cellcolor{green!25}91.42 & \cellcolor{green!25}86.28 & 88.78 & 84.79 \\
Subset B (585-class) & 88.76 & \cellcolor{green!25}85.23 & 97.37 & 90.58 & 86.13 & 88.37 & \cellcolor{green!25}87.89 \\ \hline
\end{tabular}
\end{center}
\caption{Transfer learning performance on 7 FGVC datasets by fine-tuning the Inception-v3 299 pre-trained on different source domains. Each row represents a network pre-trained on a specific source domain, and each column shows the top-1 image classification accuracy by fine-tuning different networks on a target fine-grained dataset. Relative good and poor performance on each FGVC dataset are marked by green and red, respectively. Two selected subsets based on domain similarity achieve good performance on all FGVC datasets.}
\label{tab:fgvc}
% \vspace{-2mm}
\end{table*}

\begin{figure}[t]
\centering
\includegraphics[width=\columnwidth]{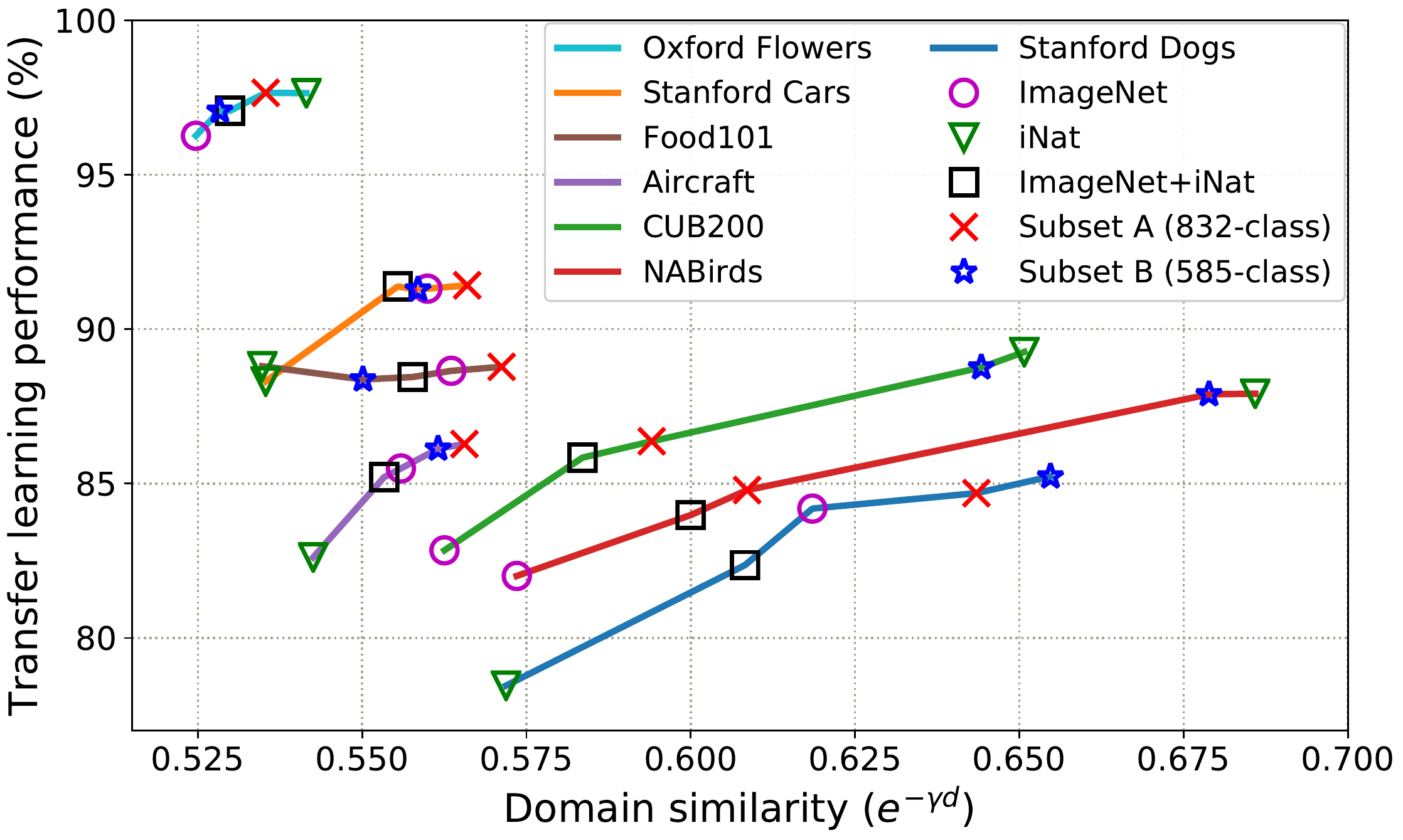}
\caption{The relationship between transfer learning performance and domain similarity between source and target domain. 
Each line represents a target FGVC dataset and each marker represents the source domain.
Better transfer learning performance can be achieved by fine-tuning the network that is pre-trained on a more similar source domain.
Two selected subsets based on our domain similarity achieve good performance on all FGVC datasets.}
\label{fig:domain-similarity}
% \vspace{-2mm}
\end{figure}

Further, in Fig.\ \ref{fig:domain-similarity}, we show the relationship between transfer learning performance and our proposed domain similarity.
We observe better transfer learning performance when fine-tuned from a more similar source domain, except Food101, on which the transfer learning performance almost stays same as domain similarity changes.
We believe this is likely due to the large number of training images in Food101 (750 training images per class).
Therefore, the target domain contains enough data thus transfer learning has very little help.
In such a scenario, our assumption on ignoring the scale of domain is no longer valid.

From Table \ref{tab:fgvc} and Fig.\ \ref{fig:domain-similarity}, we observe that the selected Subset B achieves good performance among all FGVC datasets, surpassing ImageNet pre-training by a large margin on CUB200 and NABirds.
In Table \ref{tab:fgvc_stoa}, we compare our approach with existing FGVC methods.
Results demonstrate state-of-the-art performance of the prposed transfer learning method on commonly used FGVC datasets.
Notice that since our definition of domain similarity is fast to compute, we can easily explore different ways to select a source domain.
The transfer learning performance can be directly estimated based on domain similarity, without conducting any pre-training and fine-tuning.
Prior to our work, the only options to achieve good performance on FGVC tasks are either designing better models based on ImageNet fine-tuning~\cite{bilinearcnn, kernel_pooling, multi-attention_fgvc} or augmenting the dataset by collecting more images~\cite{xu2016webly, krause2016unreasonable}.
Our work, however, provides a novel direction of using a more similar source domain to pre-train the network.
We show that with properly selected subsets in source domain, it is able to match or exceed those performance gain by simply fine-tuning off-the-shelf networks.

% Previous FGVC methods overwhelmingly focused on designing better models with ImageNet pre-trained initialization.
% Our work, however, provides a novel direction of using more similar source domains to pre-train the networks.
% We show that with a properly selected source domain, simply fine-tuning off-the-shelf networks could match or outperform the performance gain from carefully designed models.

\begin{table*}[t!]
\small
\begin{center}
\begin{tabular}{ l|c|c|c|c|c }
\hline
\textbf{Method} & \textbf{CUB200} & \textbf{Stanford Dogs} & \textbf{Stanford Cars} & \textbf{Aircrafts} & \textbf{Food101} \\ \hline
Subset B (585-class): Inception-v3 & \textbf{89.6} & 86.3 & 93.1 & 89.6 & 90.1 \\ 
Subset B (585-class): Inception-ResNet-v2 SE & 89.3 & \textbf{88.0} & \textbf{93.5} & \textbf{90.7} & \textbf{90.4} \\ \hline
Krause~\etal~\cite{krause2015fine} & 82.0 & - & 92.6 & - & - \\
Bilinear-CNN~\cite{bilinearcnn} & 84.1 & - & 91.3 & 84.1 & 82.4 \\
Compact Bilinear Pooling~\cite{cbp} & 84.3 & - & 91.2 & 84.1 & 83.2 \\
Zhang~\etal~\cite{zhang2016picking} & 84.5 & 72.0 & - & - & - \\
Low-rank Bilinear Pooling~\cite{kong2017low} & 84.2 & - & 90.9 & 87.3 & - \\
Kernel Pooling~\cite{kernel_pooling} & 86.2 & - & 92.4 & 86.9 & 85.5 \\
RA-CNN~\cite{fu2017look} & 85.3 & \textbf{87.3} & 92.5 & - & - \\
Improved Bilinear-CNN~\cite{lin2017improved} & 85.8 & - & 92.0 & 88.5 & - \\
MA-CNN~\cite{multi-attention_fgvc} & \textbf{86.5} & - & 92.8 & 89.9 & - \\
DLA~\cite{yu2017deep} & 85.1 & - & \textbf{94.1} & \textbf{92.6} & \textbf{89.7} \\ \hline
\end{tabular}
\end{center}
\caption{Comparison to existing state-of-the-art FGVC methods. 
As a convention, we use same $448 \times 448$ input size.
Since we didn't find recent proposed FGVC methods applied to Flowers-102 and NABirds, we only show comparisons on the rest of 5 datasets.
Our proposed transfer learning approach is able to achieve state-of-the-art performance on all FGVC datasets, especially on CUB200 and NABirds.
}
\label{tab:fgvc_stoa}
% \vspace{-2mm}
\end{table*}

%%%%%%%%%%%%%%%%%% Conclusions
\section{Conclusions}
\label{sec:conclusions}

In this work, we have presented a training scheme that achieves top performance on large scale iNaturalist dataset, by using higher resolution input image and fine-tuning to deal with long-tailed distribution.
We further proposed a novel way of capturing domain similarity with Earth Mover's Distance and showed better transfer learning performance can be achieved by fine-tuning from a more similar domain.
In the future, we plan to study other important factors in transfer learning beyond domain similarity.

\vspace{5pt}
\par\noindent\textbf{Acknowledgments.}
This work was supported in part by a Google Focused Research Award. We would like to thank our colleagues at Google for helpful discussions.

\newpage
%%%%%%%%%%%%%%%%%% References
{\small
\bibliographystyle{ieee}
\bibliography{ref.bib}
}

\end{document}